\documentclass[10pt, a4paper]{article}
\usepackage{lrec}

\usepackage{subcaption}
\usepackage{listings}
\usepackage{multirow}
\usepackage{amsmath}

\title{TableBank: A Benchmark Dataset for Table Detection and Recognition}

\name{Minghao Li$^1$\thanks{The paper was finished during internship at Microsoft Research Asia.}, Lei Cui$^2$, Shaohan Huang$^2$, Furu Wei$^2$, Ming Zhou$^2$, Zhoujun Li$^1$}

\address{
$^1$State Key Lab of Software Development Environment, Beihang University, Beijing, China \\
$^2$Microsoft Research Asia, Beijing, China \\
\{liminghao1630, lizj\}@buaa.edu.cn,
\{lecu, shaohanh, fuwei, mingzhou\}@microsoft.com\\}

\abstract{
We present TableBank, a new image-based table detection and recognition dataset built with novel weak supervision from Word and Latex documents on the internet. Existing research for image-based table detection and recognition usually fine-tunes pre-trained models on out-of-domain data with a few thousand human-labeled examples, which is difficult to generalize on real-world applications. With TableBank that contains 417K high quality labeled tables, we build several strong baselines using state-of-the-art models with deep neural networks. We make TableBank publicly available and hope it will empower more deep learning approaches in the table detection and recognition task. The dataset and models are  available at \url{https://github.com/doc-analysis/TableBank}. \\ \newline }

\begin{document}

\maketitleabstract

\section{Introduction}

\begin{figure*}[t]
\centering
    \begin{subfigure}[b]{0.235\textwidth}
        \includegraphics[width=\textwidth]{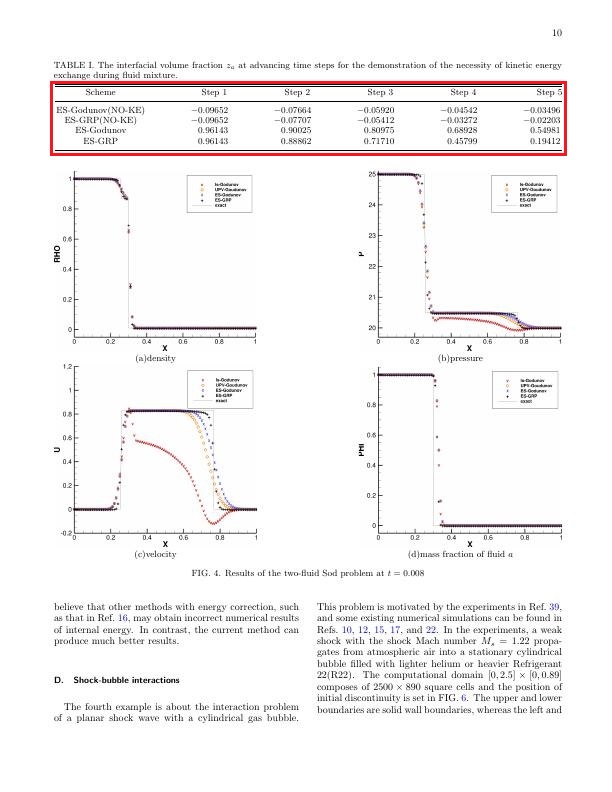}
        \caption{}
        \label{fig:1a}
    \end{subfigure}
    ~ 
    \begin{subfigure}[b]{0.235\textwidth}
        \includegraphics[width=\textwidth]{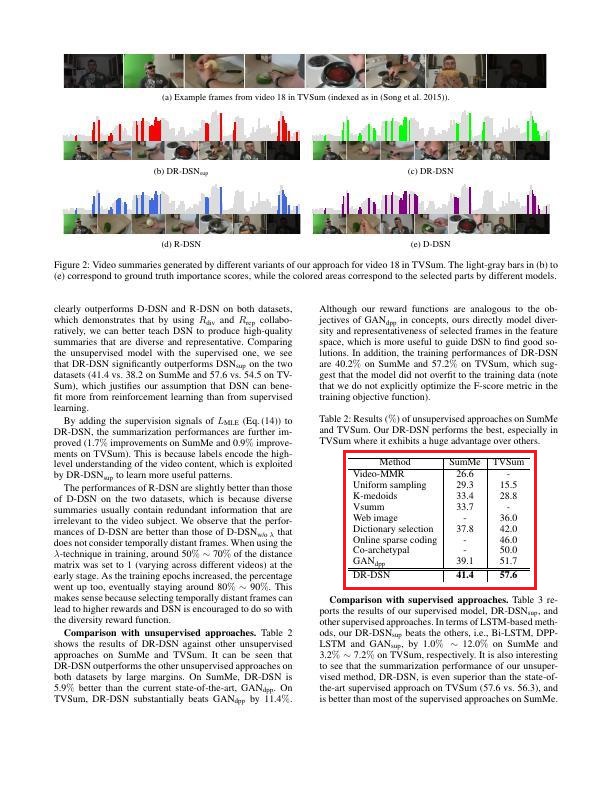}
        \caption{}
        \label{fig:1b}
    \end{subfigure}
    ~ 
    \begin{subfigure}[b]{0.235\textwidth}
        \includegraphics[width=\textwidth]{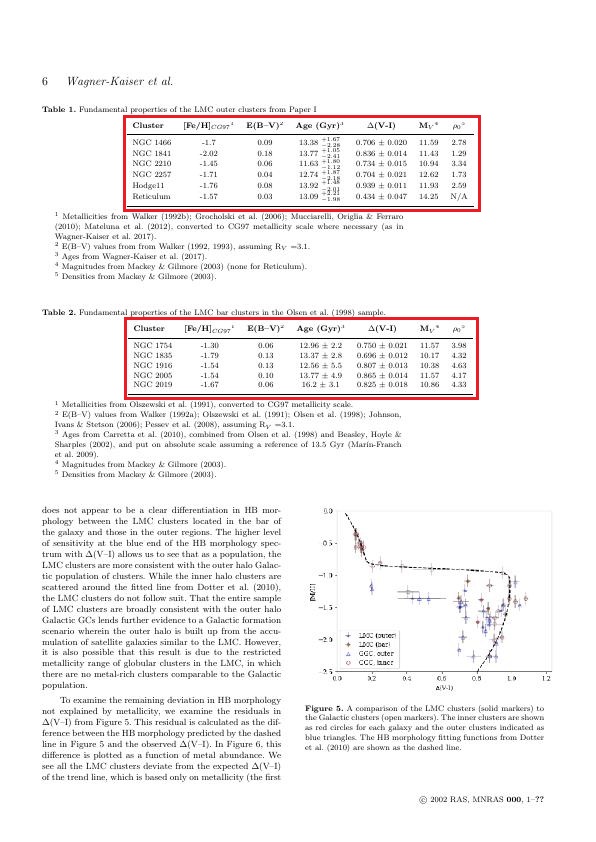}
        \caption{}
        \label{fig:1c}
    \end{subfigure}
    ~
    \begin{subfigure}[b]{0.235\textwidth}
        \includegraphics[width=\textwidth]{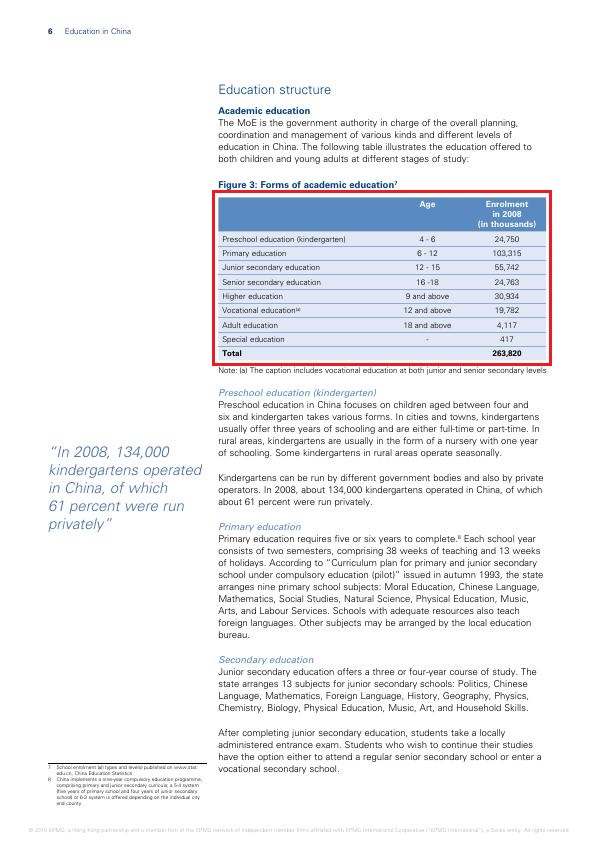}
        \caption{}
        \label{fig:1d}
    \end{subfigure}
    \caption{Tables in electronic documents on the web with different layouts and formats. Among them, Figure (a)(b)(c) come from the latex format documents while Figure (d) comes from a word format document}\label{fig:1}
\end{figure*}

Table detection and recognition is an important task in many document analysis applications as tables often present essential information in a structured way. It is a difficult problem due to varying layouts and formats of the tables as shown in Figure \ref{fig:1}. Conventional techniques for table analysis have been proposed based on the layout analysis of documents. Most of these techniques fail to generalize because they rely on handcrafted features which are not robust to layout variations. Recently, the rapid development of deep learning in computer vision has significantly boosted the data-driven image-based approaches for table analysis. The advantage of the image-based table analysis lies in its robustness to document types, making no assumption of whether scanned images of pages or natively-digital document formats. It can be applied to a wide variety of document types to extract tables including PDF, HTML, PowerPoint as well as their scanned copies. Although some document types may have structured tabular data, we still need a general approach to detect tables on different kinds of documents. Therefore, large-scale end-to-end deep learning models make it feasible for achieving better performance.


Existing deep learning based table analysis models usually fine-tune pre-trained object detection models with several thousand human-labeled training instances, while still difficult to scale up in real-world applications. For instance, we found that models trained on data similar to Figure~\ref{fig:1a}~\ref{fig:1b}~\ref{fig:1c} would not perform well on Figure~\ref{fig:1d} because the table layouts and colors are so different. Therefore, enlarging the training data should be the only way to build open-domain table analysis models with deep learning. Deep learning models are massively more complex than most traditional models, where many standard deep learning models today have hundreds of millions of free parameters and require considerably more labeled training data. In practice, the cost and inflexibility of hand-labeling such training sets is the key bottleneck to actually deploying deep learning models. It is well known that ImageNet~\cite{ILSVRC15} and COCO~\cite{DBLP:journals/corr/LinMBHPRDZ14} are two popular image classification and object detection datasets that are built in a crowdsourcing way, while they are both expensive and time-consuming to create, taking months or years for large benchmark sets. Fortunately, there exists a large number of digital documents on the internet such as Word and Latex source files. It is instrumental if some weak supervision can be applied to these online documents for labeling tables.

To address the need for a standard open-domain table benchmark dataset, we propose a novel weak supervision approach to automatically create the TableBank, which is orders of magnitude larger than existing human-labeled datasets for table analysis. Distinct from the traditional weakly supervised training set, our approach can obtain not only large scale but also high quality training data. Nowadays, there are a great number of electronic documents on the web such as Microsoft Word (.docx) and Latex (.tex) files. These online documents contain mark-up tags for tables in their source code by nature. Intuitively, we can manipulate these source code by adding bounding boxes using the mark-up language within each document. For Word documents, the internal Office XML code can be modified where the borderline of each table is identified. For Latex documents, the tex code can be also modified where bounding boxes of tables are recognized. In this way, high quality labeled data is created for a variety of domains such as business documents, official filings, research papers, etc, which is tremendously beneficial for large-scale table analysis tasks.

The TableBank dataset totally consists of 417,234 high quality labeled tables as well as their original documents in a variety of domains. To verify the effectiveness of TableBank, we build several strong baselines using state-of-the-art models with end-to-end deep neural networks. The table detection model is based on the Faster R-CNN~\cite{DBLP:journals/corr/RenHG015} architecture with different settings. The table structure recognition model is based on the encoder-decoder framework for image-to-text. The experiment results show that the layout and format variation has a great impact on the accuracy of table analysis tasks. In addition, models trained in one specific domain do not perform well in the other. This suggests that there is plenty of room for advancement in modeling and learning on the TableBank dataset.


\section{Existing Datasets}

We introduce some existing publicly available datasets that we compare as the baselines:

\noindent\textbf{ICDAR 2013 Table Competition.} The ICDAR 2013 Table Competition dataset~\cite{6628853} contains 128 examples in natively-digital document format, which are from the European Union and US Government.

\noindent\textbf{UNLV Table Dataset.} The UNLV Table Dataset~\cite{Shahab:2010:OAT:1815330.1815345} contains 427 examples in scanned image format, which are from a variety of sources including Magazines, Newspapers, Business Letter, Annual Report, etc.

\noindent\textbf{Marmot Dataset.} The Marmot Dataset\footnote{\small \url{http://www.icst.pku.edu.cn/cpdp/data/marmot_data.htm}} contains 2,000 pages in PDF format, where most of the examples are from research papers.

\noindent\textbf{DeepFigures Dataset.} The DeepFigures Dataset~\cite{DBLP:journals/corr/abs-1804-02445} includes documents with tables and figures from arXiv.com and PubMed database. The DeepFigures Dataset focuses on the large scale table/figure detection task while it does not contain the table structure recognition dataset.

\section{Data Collection}

Basically, we create the TableBank dataset using two different file types: Word documents and Latex documents. Both file types contain mark-up tags for tables in their source code by nature. Next, we introduce the details in three steps: document acquisition, creating table detection dataset and table structure recognition dataset.

\subsection{Document Acquisition}

We crawl Word documents from the internet. The documents are all in `.docx' format since we can edit the internal Office XML code to add bounding boxes. Since we do not filter the document language, the Word documents contain English, Chinese, Japanese, Arabic, and other languages. This makes the dataset more diverse and robust to real applications.

Latex documents are different from Word documents because they need other resources to be compiled into PDF files. Therefore, we cannot only crawl the `.tex' file from the internet. Instead, we use documents from the largest pre-print database arXiv.org as well as their source code. We download the Latex source code from 2014 to 2018 through bulk data access in arXiv. The language of the Latex documents is mainly English.

\subsection{Table Detection}

\begin{figure}[h]
\centering
\includegraphics[width=0.45\textwidth]{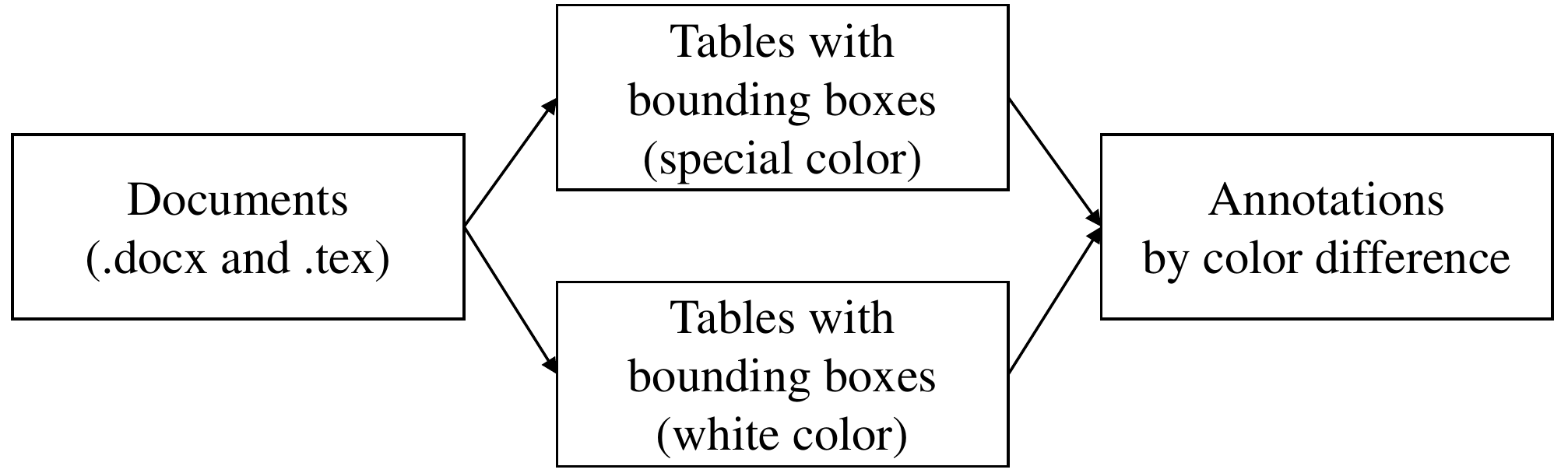}
\caption{Data processing pipeline}\label{fig:2}
\end{figure}

\begin{figure*}[ht]
\centering
\includegraphics[width=0.85\textwidth]{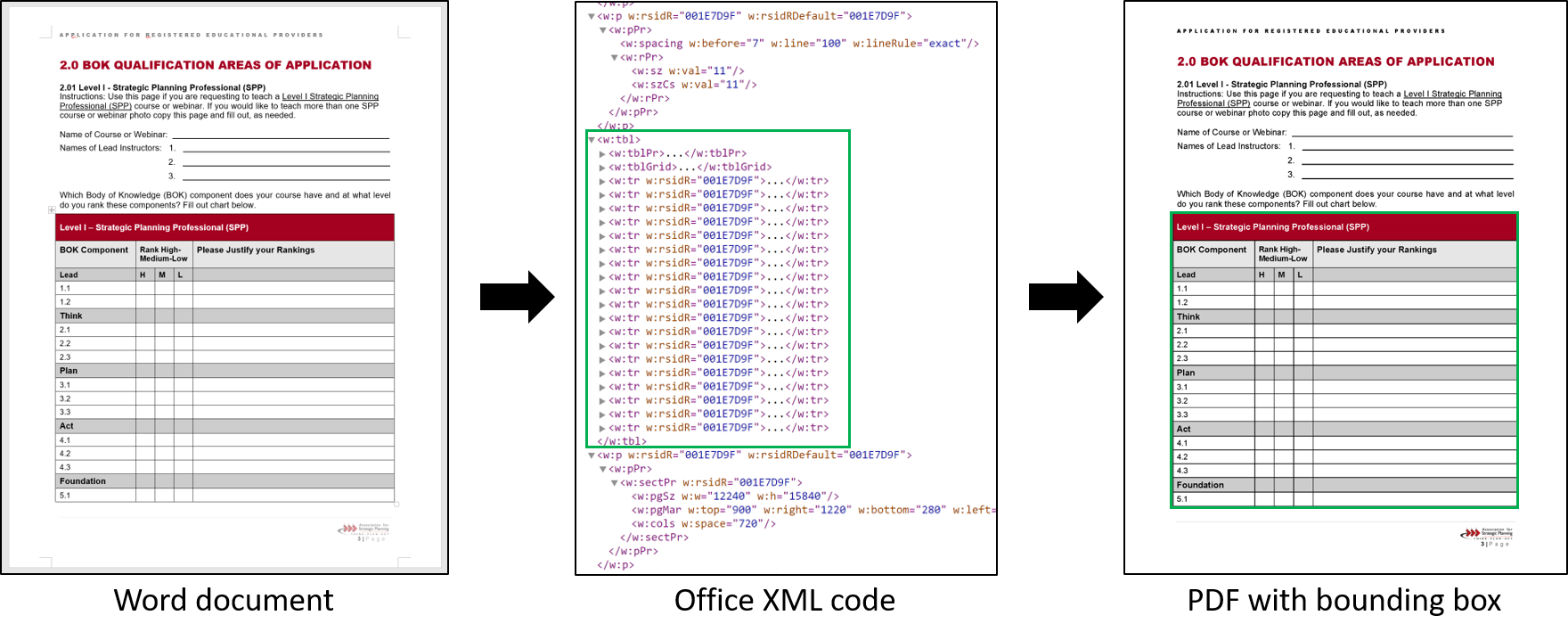}
\caption{The tables can be identified and labeled from ``$<$w:tbl$>$'' and ``$<$/w:tbl$>$'' tags in the Office XML code}\label{fig:3}
\end{figure*}

Intuitively, we can manipulate the source code by adding bounding boxes using the mark-up language within each document. The processing pipeline is shown in Figure~\ref{fig:2}. For Word documents, we add the bounding boxes to tables through editing the internal Office XML code within each document. Actually, each `.docx' file is a compressed archive file. There exists a `document.xml' file in the decompressed folder of the `.docx' file. In the XML file, the code snippets between the tags `$<$w:tbl$>$' and `$<$/w:tbl$>$' usually represent a table in the Word document, which is shown in Figure~\ref{fig:3}. We modified the code snippets in the XML file so that the table borders can be changed into a distinguishable color compared to other parts of the document. Figure~\ref{fig:3} shows that we have added a green bounding box in the PDF file where the table is perfectly identified. Finally, we get PDF pages from Word documents that contain at least one table on each page.

For Latex documents, we use a special command in the Tex grammar `fcolorbox' to add the bounding box to tables. Typically, the tables in Latex documents are usually in the format as follows:
\begin{center}
\begin{tabular}{|c|}
\hline
\begin{lstlisting}
\begin{table}[]
    \centering
    \begin{tabular}{}
         ...
    \end{tabular}
\end{table}
\end{lstlisting}
\\\hline
\end{tabular}
\end{center}
We insert the `fcolorbox' command to the table's code as follows and re-compile the Latex documents. Meanwhile, we also define a special color so that the border is distinguishable. The overall process is similar to Word documents. Finally, we get PDF pages from Latex documents that contain at least one table on each page.
\begin{center}
\begin{tabular}{|c|}
\hline
\begin{lstlisting}
\begin{table}[]
    \centering
    \setlength{\fboxsep}{1pt}
    \fcolorbox{bordercolor}{white}{
    \begin{tabular}{}
         ...
    \end{tabular}}
\end{table}
\end{lstlisting}
\\\hline
\end{tabular}
\end{center}


\begin{figure*}[t]
\centering
    \begin{subfigure}[b]{0.235\textwidth}
        \includegraphics[width=\textwidth]{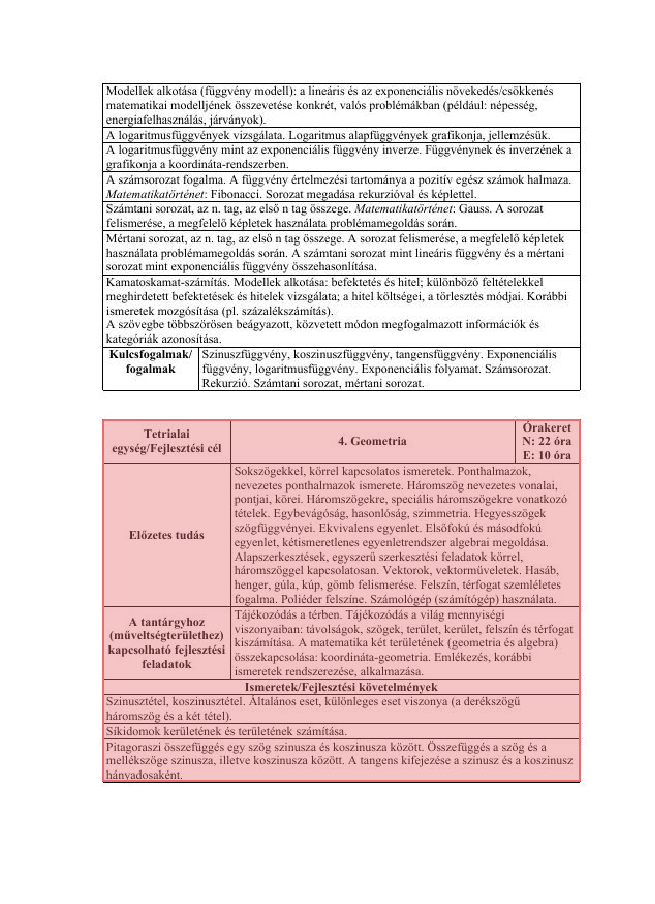}
        \caption{Missed}
    \end{subfigure}
    ~ 
    \begin{subfigure}[b]{0.235\textwidth}
        \includegraphics[width=\textwidth]{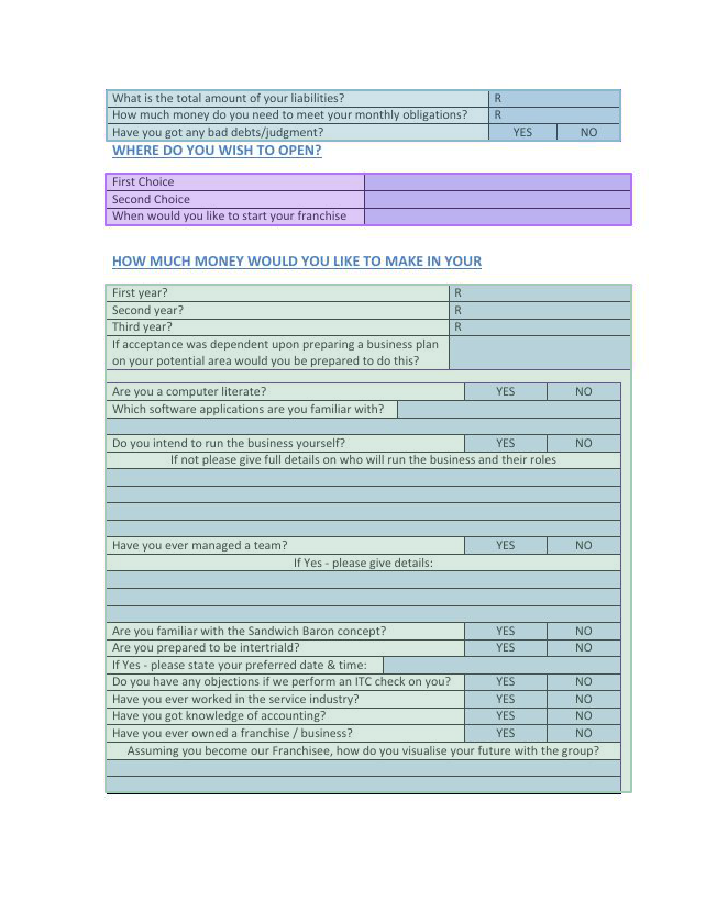}
        \caption{Mixed}
    \end{subfigure}
    ~ 
    \begin{subfigure}[b]{0.235\textwidth}
        \includegraphics[width=\textwidth]{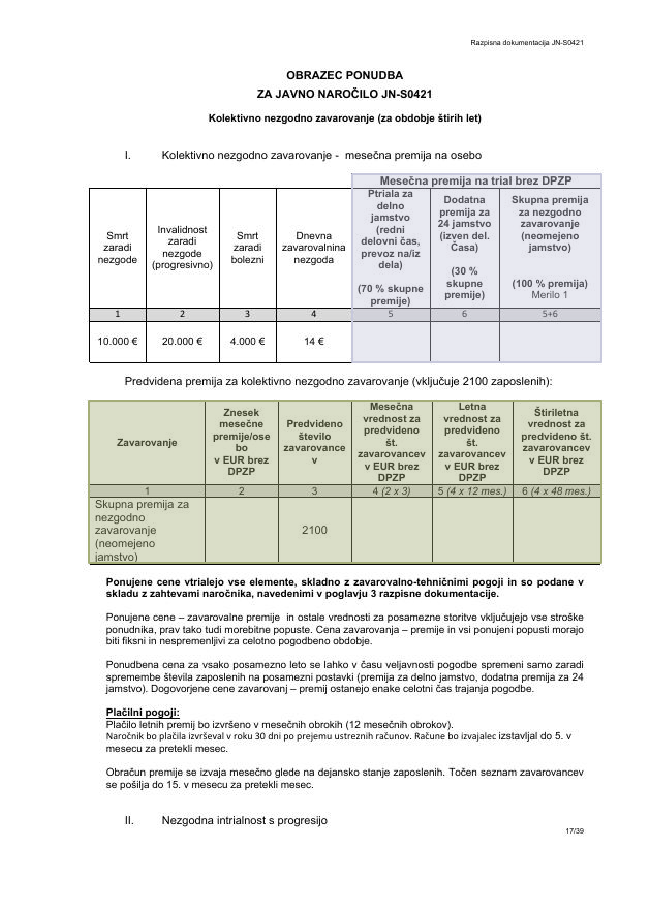}
        \caption{Incomplete}
    \end{subfigure}
    ~
    \begin{subfigure}[b]{0.235\textwidth}
        \includegraphics[width=\textwidth]{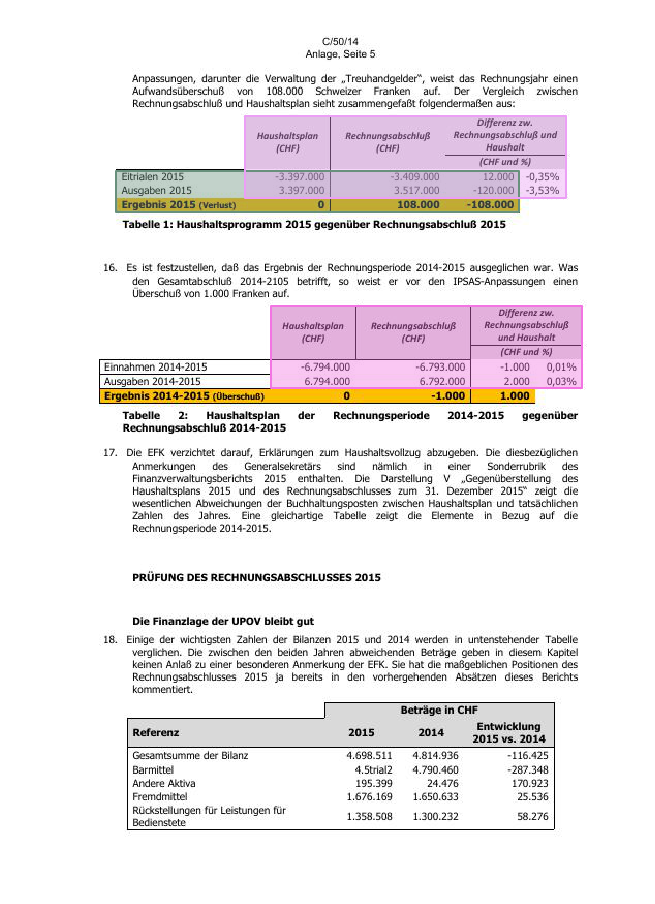}
        \caption{Overlaped}
    \end{subfigure}
    \caption{Figure (a): Missed a table; Figure (b): Two tables were annotated as one; Figure (c): One of the annotations is incomplete; Figure (c): Some annotations are overlaped. }
    \label{errors}
\end{figure*}

To obtain ground truth labels, we extract the annotations from the generated PDF documents. For each page in the annotated PDF documents, we also add the bounding boxes to the tables in the original page with the white color so that the two pages align in the same position. Then, we simply compare two pages in the pixel-level so that we can find the different pixels and get the upper left point of the tables as well as the width and height parameters. In this way, we totally create 417,234 labeled tables from the crawled documents. We randomly sample 1,000 examples from the dataset and manually check the bounding boxes of tables. We observe that only 5 of them are incorrectly labeled, which demonstrates the high quality of this dataset. We give several typical weak supervised labeled errors in Figure~\ref{errors}.

\subsection{Table Structure Recognition}

\begin{figure}[ht]
\centering
\includegraphics[width=0.45\textwidth]{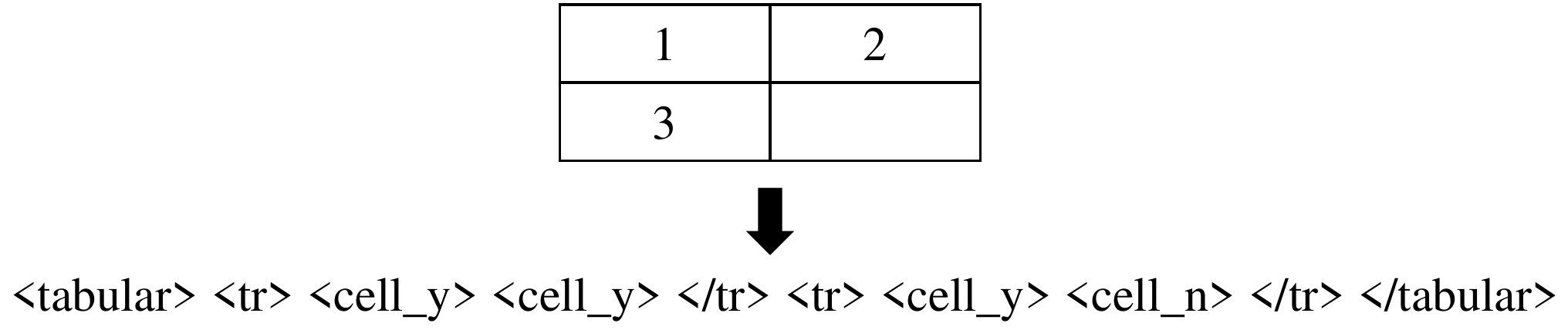}
\caption{A Table-to-HTML example where `$<$cell\_y$>$' denotes cells with content while `$<$cell\_n$>$' represents cells without content}\label{fig:4}
\end{figure}

Table structure recognition aims to identify the row and column layout structure for the tables especially in non-digital document formats such as scanned images. Existing table structure recognition models usually identify the layout information as well as the textual content of the cells, while textual content recognition is not the focus of this work. Therefore, we define the task as follows: given a table in the image format, generating an HTML tag sequence that represents the arrangement of rows and columns as well as the type of table cells. In this way, we can automatically create the structure recognition dataset from the source code of Word and Latex documents. For Word documents, we simply transform the original XML information from documents into the HTML tag sequence. For Latex documents, we first use the LaTeXML toolkit\footnote{\url{https://dlmf.nist.gov/LaTeXML/}} to generate XML from Latex, then transform the XML into HTML. A simple example is shown in Figure~\ref{fig:4}, where we use `$<$cell\_y$>$' to denotes the cells with content and `$<$cell\_n$>$' to represent the cells without content. After filtering the noise, we create a total of 145,463 training instances from the Word and Latex documents.

In the process of inferring, we get the row and column structures from table structure recognition results, as well as the content and bounding boxes of all the closely arranged text blocks from OCR results. Assuming a table contains $N$ rows, row gaps between OCR bounding boxes are detected by a heuristic algorithm to split the table content blocks into $N$ groups. Finally, we fill the blocks in the `$<$cell\_y$>$' tag of $N$ lines with the order from left to right.

For example, the table in Figure~\ref{fig:4} has two rows according to table structure recognition result, e.g. the HTML tag sequence. According to the row gap between text block ``1'' and text block ``3'', we divide them into two groups: $\langle$1, 2$\rangle$ and $\langle$3$\rangle$. The first group corresponds the first row containing two `$<$cell\_y$>$' tags, and the second group corresponds the second row containing a `$<$cell\_y$>$' tag and a `$<$cell\_n$>$' tag.

\begin{figure}[ht]
\centering
\includegraphics[width=0.45\textwidth]{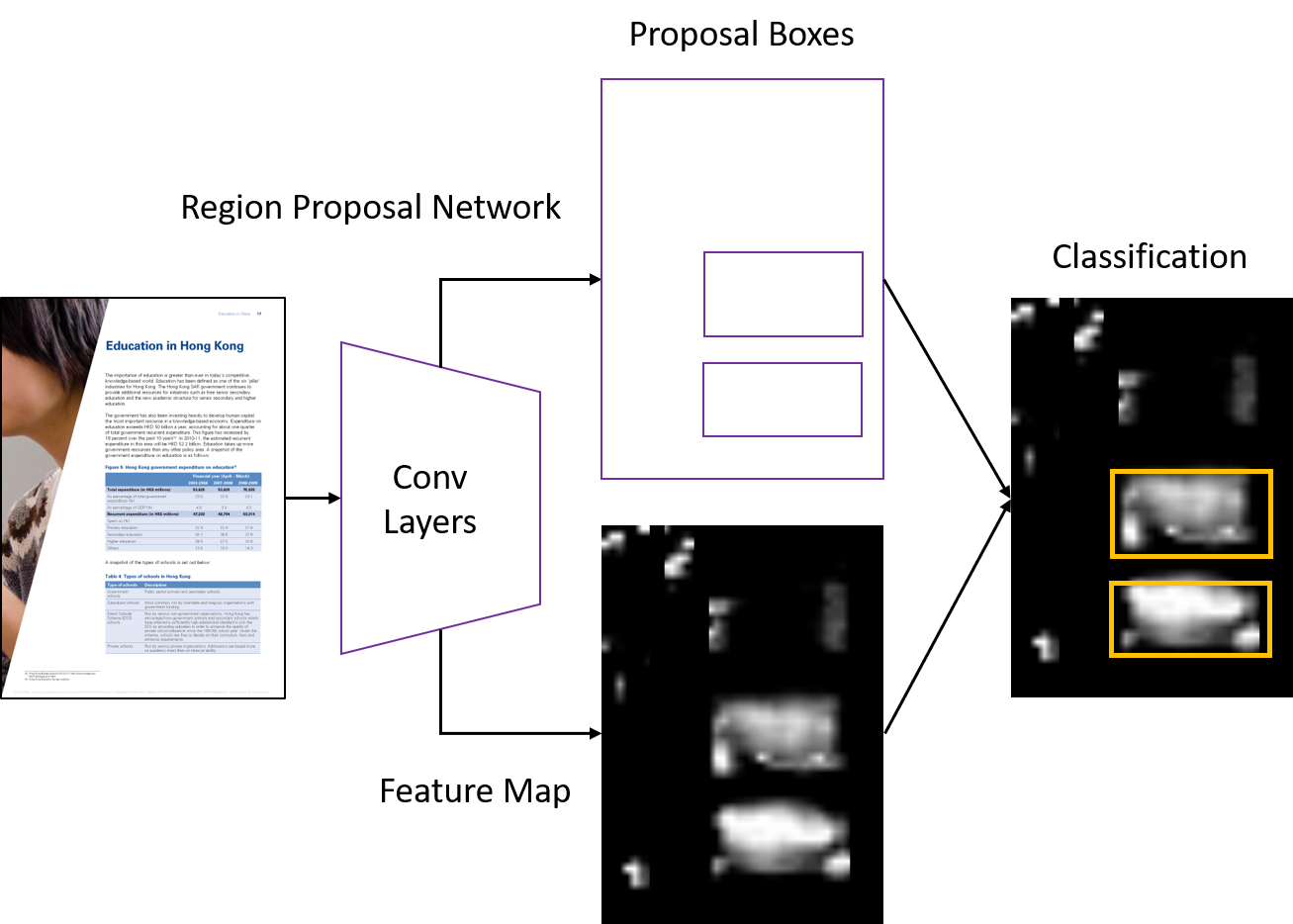}
\caption{The Faster R-CNN model for table detection}\label{fig:5}
\end{figure}

\section{Baseline}

\subsection{Table Detection}

We use the Faster R-CNN model as the baseline. Due to the success of ImageNet and COCO competition, the Faster R-CNN model has been the de facto implementation that is widespread in the computer vision area. In 2013, the R-CNN model ~\cite{DBLP:journals/corr/GirshickDDM13} was first proposed to solve the object detection problem. After that,~\cite{DBLP:journals/corr/Girshick15} also proposed the Fast R-CNN model that improves training and testing speed while also increasing detection accuracy. Both models use selective search to find out the region proposals, while selective search is a slow and time-consuming process affecting the performance of the network. Distinct from two previous methods, the Faster R-CNN method introduces a Region Proposal Network (RPN) that shares full-image convolutional features with the detection network, thus enabling nearly cost-free region proposals. Furthermore, the model merges RPN and Fast R-CNN into a single network by sharing their convolutional features so that the network can be trained in an end-to-end way. The overall architecture of Faster R-CNN in shown in Figure~\ref{fig:5}.

\subsection{Table Structure Recognition}

\begin{figure}[ht]
\centering
\includegraphics[width=0.45\textwidth]{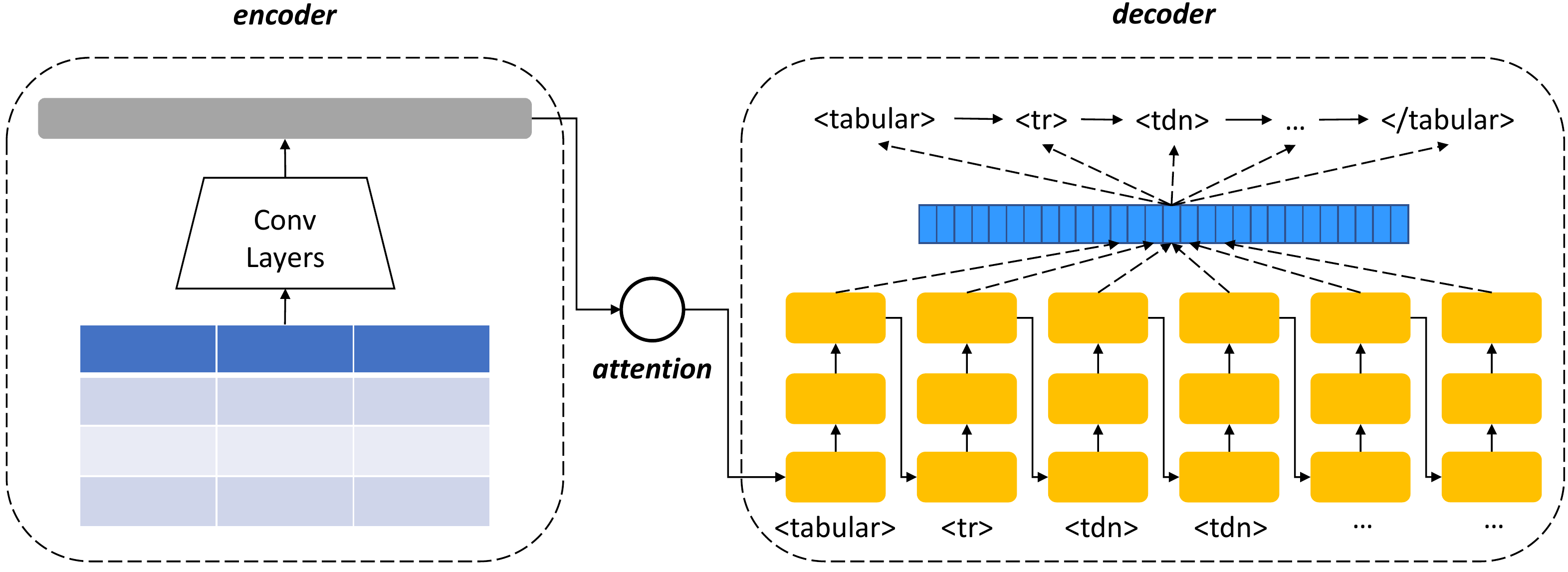}
\caption{Image-to-Text model for table structure recognition}\label{fig:6}
\end{figure}

We leverage the image-to-text model as the baseline. The image-to-text model has been widely used in image captioning, video description, and many other applications. A typical image-to-text model includes an encoder for the image input and a decoder for the text output. In this work, we use the image-to-markup model~\cite{DBLP:journals/corr/DengKR16} as the baseline to train models on the TableBank dataset. The overall architecture of the image-to-text model is shown in Figure~\ref{fig:6}.

\section{Experiment}

\subsection{Data and Metrics}

The statistics of TableBank is shown in Table~\ref{tab:1}. To evaluate table detection, we sample 2,000 document images from Word and Latex documents respectively, where 1,000 images for validation and 1,000 images for testing. Each sampled image contains at least one table. Meanwhile, we also evaluate our model on the ICDAR 2013 dataset to verify the effectiveness of TableBank. To evaluate table structure recognition, we sample 500 tables each for validation and testing from Word documents and Latex documents respectively. The size of the training set for table detection and table structure recognition is 415,234 and 144,463 respectively. The entire training and testing data have been made available to the public. For table detection, we calculate the precision, recall, and F1 in the same way as in~\cite{8270062}, where the metrics for all documents are computed by summing up the area of overlap, prediction and ground truth. The definition is defined as follows:
\begin{small}
\begin{gather*}
Precision=\frac{\text{Area of Ground truth regions in Detected regions}}{\text{Area of all Detected table regions}},\notag \\
Recall=\frac{\text{Area of Ground truth regions in Detected regions}}{\text{Area of all Ground truth table regions}},\notag \\
F1\text{ }Score=\frac{\text{2} \times \text{Precision} \times \text{Recall}}{\text{Precision + Recall}}\notag.
\end{gather*}  
\end{small}

For table structure recognition, we use the 4-gram BLEU score as the evaluation metric with a single reference. The 4-gram BLEU score has the advantages of fast calculation speed, low cost, and being easy to understand. 

\begin{table}[ht]
\centering
\scalebox{0.84}{
\begin{tabular}{|c|c|c|c|}
\hline \bf Task & \bf Word & \bf Latex & \bf Word+Latex  \\ \hline\hline
Table detection  & 163,417  & 253,817 &   417,234  \\ \hline
Table structure recognition  & 56,866  & 88,597 &  145,463 \\ \hline
\end{tabular}}
\caption{Statistics of TableBank} 
\label{tab:1}
\end{table}


\subsection{Settings}

For table detection, we use the open source framework Detectron~\cite{Detectron2018} to train models on the TableBank. Detectron is a high quality and high-performance codebase for object detection research, which supports many state-of-the-art algorithms. It is written in Python and powered by the Caffe2 deep learning framework. In this task, we use the Faster R-CNN algorithm with the ResNeXt~\cite{DBLP:journals/corr/XieGDTH16} as the backbone network architecture, where the parameters are pre-trained on the ImageNet dataset. All baselines are trained using 4$\times$P100 NVIDIA GPUs using data parallel sync SGD with a minibatch size of 16 images. For other parameters, we use the default values in Detectron. During testing, the confidence threshold of generating bounding boxes is set to 90\%.

For table structure recognition, we use the open source framework OpenNMT~\cite{opennmt} to train the image-to-text model. OpenNMT is an open-source ecosystem for neural machine translation and neural sequence learning, which supports many encoder-decoder frameworks. In this task, we train our model using the image-to-text method in OpenNMT. The model is also trained using 4$\times$P100 NVIDIA GPUs with the learning rate of 0.1 and batch size of 24. In this task, the vocabulary size of the output space is small, including $<$tabular$>$, $<$/tabular$>$, $<$thead$>$, $<$/thead$>$, $<$tbody$>$, $<$/tbody$>$, $<$tr$>$, $<$/tr$>$, $<$td$>$, $<$/td$>$, $<$cell\_y$>$, $<$cell\_n$>$. For other parameters, we use the default values in OpenNMT.

\subsection{Results}

\begin{table*}[ht]
\centering
\small
\begin{tabular}{|c|c|c|c|c|c|c|c|c|c|}
\hline \multirow{2}{1.5cm}{\bf Models} & \multicolumn{3}{c|}{\bf Word}  & \multicolumn{3}{c|}{\bf Latex} & \multicolumn{3}{c|}{\bf Word+Latex} \\ \cline{2-10}
 & \multicolumn{1}{c|}{\bf Precision} & \multicolumn{1}{c|}{\bf Recall} & \multicolumn{1}{c|}{\bf F1} & \multicolumn{1}{c|}{\bf Precision} & \multicolumn{1}{c|}{\bf Recall} & \multicolumn{1}{c|}{\bf F1}  & \multicolumn{1}{c|}{\bf Precision} & \multicolumn{1}{c|}{\bf Recall} & \multicolumn{1}{c|}{\bf F1} \\ \hline\hline
ResNeXt-101 (Word)  &  0.9496 & 0.8388  & 0.8908 & 0.9902 & 0.5948 & 0.7432 & 0.9594 & 0.7607 & 0.8486 \\ \hline
ResNeXt-152 (Word)  & 0.9530 & 0.8829 & \bf 0.9166 & 0.9808 & 0.6890 & 0.8094 & 0.9603 & 0.8209 & 0.8851 \\ \hline
ResNeXt-101 (Latex)  & 0.8288  & 0.9395 & 0.8807 & 0.9854 & 0.9760 & 0.9807 & 0.8744 & 0.9512 & 0.9112 \\ \hline
ResNeXt-152 (Latex)  & 0.8259 & 0.9562 & 0.8863 & 0.9867 & 0.9754  & \bf 0.9810 & 0.8720 & 0.9624 & 0.9149 \\\hline
ResNeXt-101 (Word+Latex) & 0.9557 & 0.8403 & 0.8943 & 0.9886 & 0.9694 & 0.9789 & 0.9670 & 0.8817 & 0.9224\\ \hline
ResNeXt-152 (Word+Latex) & 0.9540 & 0.8639 & 0.9067 & 0.9885 & 0.9732 & 0.9808 & 0.9657 & 0.8989 & \bf 0.9311 \\\hline
\end{tabular}
\caption{Evaluation results on Word and Latex datasets with ResNeXt-\{101,152\} as the backbone networks} 
\label{tab:2}
\end{table*}

The evaluation results of table detection models are shown in Table~\ref{tab:2}. We observe that models perform well on the same domain. For instance, the ResNeXt-152 model trained with Word documents achieves an F1 score of 0.9166 on the Word dataset, which is much higher than the F1 score (0.8094) on Latex documents. Meanwhile, the ResNeXt-152 model trained with Latex documents achieves an F1 score of 0.9810 on the Latex dataset, which is also much higher than testing on the Word documents (0.8863). This indicates that the tables from different types of documents have different visual appearance. Therefore, we cannot simply rely on transfer learning techniques to obtain good table detection models with small scale training data. When combining training data with Word and Latex documents, the accuracy of larger models is comparable to models trained on the same domain, while it performs better on the Word+Latex dataset. This verifies that model trained with larger data generalizes better on different domains, which illustrates the importance of creating a larger benchmark dataset. 

\begin{table}[ht]
\centering
\scalebox{0.83}{
\small
\begin{tabular}{|c|c|c|c|}
\hline \bf Models & \bf Precision &\bf Recall & \bf F1   \\\hline\hline
ICDAR 2013 (train)  & 0.9748 & 0.7997 & 0.8786  \\ \hline
UNLV  & 0.9185 & 0.9639 & 0.9406 \\ \hline
Marmot  & 0.7692 & 0.9844 & 0.8636 \\ \hline
DeepFigures  & 0.8527 & 0.9348 & 0.8918    \\ \hline
TableBank (ResNeXt-152, Word)  & 0.9725 & 0.8528 & 0.9087   \\ \hline
TableBank (ResNeXt-152, Latex)  & \bf 0.9658 & \bf 0.9594 & \bf 0.9625  \\ \hline
TableBank (ResNeXt-152, Word + Latex)  & 0.9635 & 0.9039 & 0.9328 \\ \hline\hline
Tesseract   & 0.9439 & 0.7144 & 0.8133  \\ \hline
Camelot & 0.9785 & 0.6856 & 0.8063 \\\hline
\end{tabular}
}
\caption{Evaluation results on ICDAR 2013 dataset} 
\label{tab:compare}
\end{table}

In addition, we also evaluate our models on the ICDAR 2013 table competition dataset shown in Table~\ref{tab:compare}. Among all the models trained with TableBank, the Latex ResNeXt-152 model achieves the best F1 score of 0.9625. Furthermore, we compare our model with models trained with ICDAR, UNLV, Marmot, and DeepFigures, which demonstrates the great potential of our TableBank dataset. This also illustrates that we not only need large scale training data but also high quality data. We also compare with the open source Tesseract\footnote{\url{https://github.com/tesseract-ocr/tesseract}} and Camelot\footnote{\url{https://github.com/socialcopsdev/camelot}} toolkits, which are based on line information. The results show that image-based deep learning models are significantly better than conventional approaches.

\begin{table}[ht]
\centering
\scalebox{0.86}{
\begin{tabular}{|c|c|c|c|}
\hline \bf Models & \bf Word & \bf Latex & \bf Word+Latex \\ \hline\hline
Image-to-Text (Word)  & \bf 0.7507 & 0.6733 & 0.7138  \\ \hline
Image-to-Text (Latex)  & 0.4048 & \bf 0.7653 & 0.5818  \\ \hline
Image-to-Text (Word+Latex)  & 0.7121 & 0.7647 & \bf 0.7382 \\ \hline
\end{tabular}
}
\caption{Evaluation results (BLEU) for image-to-text models on Word and Latex datasets} 
\label{tab:3}
\end{table}

The evaluation results of table structure recognition are shown in Table~\ref{tab:3}. We observe that the image-to-text models also perform better on the same domain. The model trained on Word documents performs much better on the Word test set than the Latex test set and vice versa. Similarly, the model accuracy of the Word+Latex model is comparable to other models on Word and Latex domains and better on the mixed-domain dataset. This demonstrates that the mixed-domain model might generalize better in real-world applications.

        


\subsection{Analysis}

\begin{figure*}[t]
\centering
    \begin{subfigure}[b]{0.25\textwidth}
        \includegraphics[width=\textwidth]{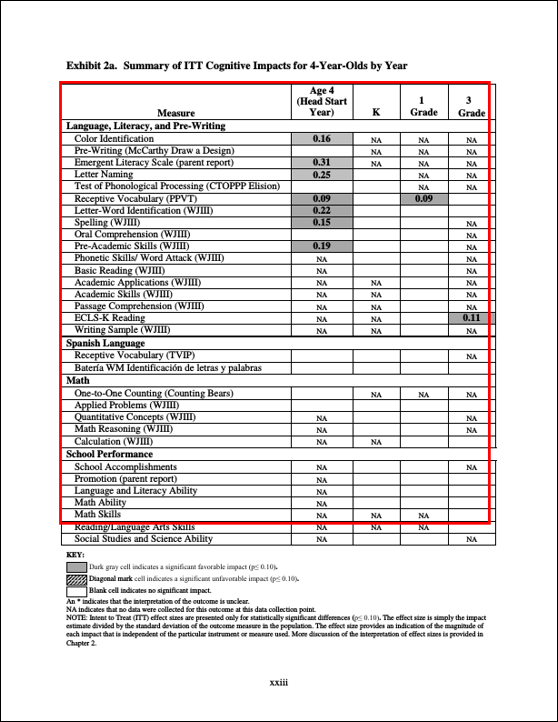}
        \caption{}
        \label{fig:7a}
    \end{subfigure}
    ~ 
    \begin{subfigure}[b]{0.25\textwidth}
        \includegraphics[width=\textwidth]{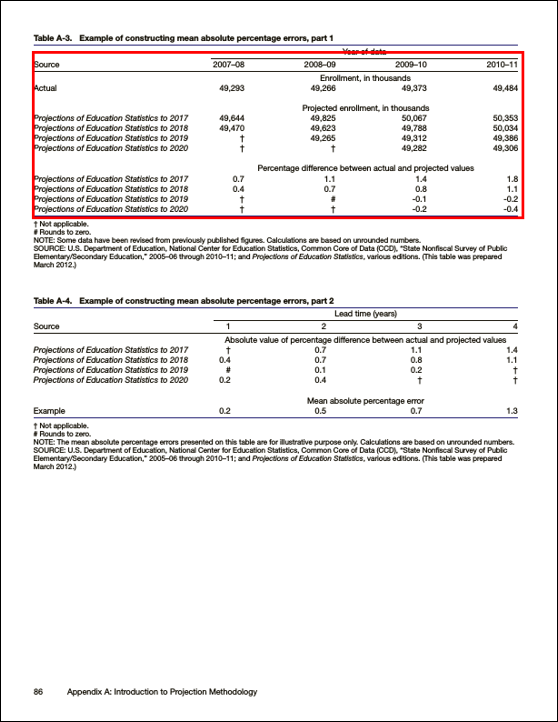}
        \caption{}
        \label{fig:7b}
    \end{subfigure}
    ~ 
    \begin{subfigure}[b]{0.25\textwidth}
        \includegraphics[width=\textwidth]{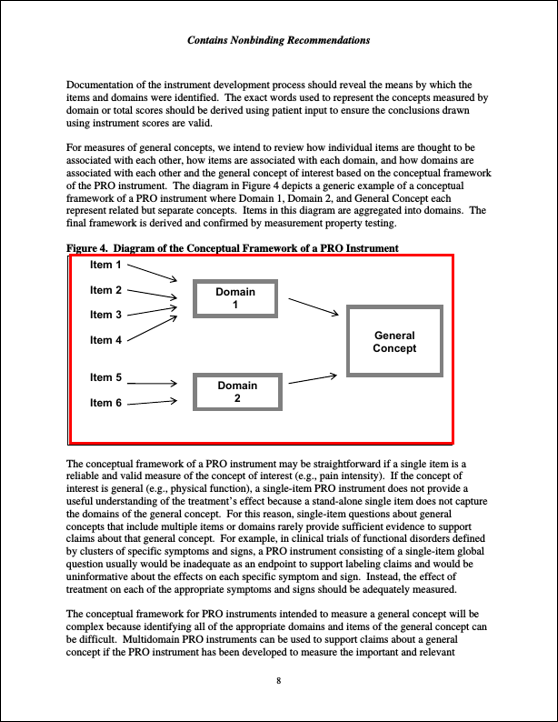}
        \caption{}
        \label{fig:7c}
    \end{subfigure}
    \caption{Table detection examples with (a) partial-detection, (b) un-detection and (c) mis-detection}\label{fig:7}
\end{figure*}

For table detection, we sample some incorrect examples from the evaluation data of Word and Latex documents for the case study. Figure~\ref{fig:7} gives three typical errors of detection results. The first error type is \textbf{partial-detection}, where only part of the tables can be identified and some information is missing. The second error type is \textbf{un-detection}, where some tables in the documents cannot be identified. The third error type is \textbf{mis-detection}, where figures and text blocks in the documents are sometimes identified as tables. Taking the ResNeXt-152 model for Word+Latex as an example, the number of un-detected tables is 164. Compared with ground truth tables (2,525), the un-detection rate is 6.5\%. Meanwhile, the number of mis-detected tables is 86 compared with the total predicted tables being 2,450. Therefore, the mis-detection rate is 3.5\%. Finally, the number of partial-detected tables is 57, leading to a partial-detection rate of 2.3\%. This illustrates that there is plenty of room to improve the accuracy of the detection models, especially for un-detection and mis-detection cases.

We observed that some experiments show inferior accuracy during cross-domain testing. The precision of the Latex model is not satisfactory when tested on the Word dataset and the Word+Latex dataset. Meanwhile, the recall of the Word model is poor on all the test dataset, and the recall of the mixed model is also under-performed on the Word dataset and Word+Latex dataset. It shows that there is still a lot of room to improve on the TableBank dataset, as well as some bad cases to be investigated especially under cross-domain settings.

For table structure recognition, we observe that the model accuracy reduces as the length of output becomes larger. Taking the image-to-text model for Word+Latex as an example, the number of exact match between the output and ground truth is shown in Table~\ref{tab:4}. We can see that the ratio of exact match is around 50\% for the HTML sequences that are less than 40 tokens. As the number of tokens becomes larger, the ratio reduces dramatically to 8.6\%, indicating that it is more difficult to recognize big and complex tables. In general, the model totally generates the correct output for 338 tables. 
In order to avoid the influence of the length distribution deviation in the training data, we also count the length distribution in the training data in Table~\ref{tab:5}.
We believe enlarging the training data will further improve the current model especially for tables with complex row and column layouts, which will be our next-step effort.

\begin{table}[ht]
\centering
\scalebox{0.83}{
\begin{tabular}{|c|c|c|c|c|c|c|}
\hline \bf Length & \bf 0-20 & \bf 21-40 & \bf 41-60 & \bf 61-80 & \bf $>$80 & \bf All \\ \hline\hline
$\#$Total  & 32 & 293 & 252 & 145 & 278 & 1,000 \\ \hline
$\#$Exact match & 15 & 169 & 102 & 28 & 24 & 338 \\ \hline
Ratio & 0.469 & 0.577 & 0.405 & 0.193 & 0.086 & 0.338\\ \hline
\end{tabular}}
\caption{Number of exact match between the generated HTML tag sequence and ground truth sequence} 
\label{tab:4}
\end{table}

\begin{table}[ht]
\centering
\scalebox{0.83}{
\begin{tabular}{|c|c|c|c|c|c|c|}
\hline \bf Length & \bf 0-20 & \bf 21-40 & \bf 41-60 & \bf 61-80 & \bf $>$80 & \bf All \\ \hline\hline
$\#$Total  & 4,027 & 44,811 & 36,059 & 19,757 & 40,809 & 145,463 \\ \hline
Ratio & 0.028 & 0.308 & 0.248 & 0.136 & 0.281 & 1.000\\ \hline
\end{tabular}}
\caption{Length distribution in training data} 
\label{tab:5}
\end{table}

\section{Related Work}

\subsection{Table Detection}

Table detection aims to locate tables using bounding boxes in a document. The research of table detection dates back to the early 1990s.~\cite{395625} proposed a rule-based approach that leverages the text block arrangement and ruled line position to detect table structures. At the same time,~\cite{395683} designed a structural table detection method based on horizontal and vertical lines, as well as the item blocks. Following these works, there is a great deal of research work~\cite{601964,green1995recognition,Tupaj96extractingtabular,medium-independent,Gatos:2005:ATD:2081228.2081298,35652} focus on improving rule-based systems. Although these methods perform well on some documents, they require extensive human efforts to figure out better rules, while sometimes failing to generalize to documents from other sources. Therefore, it is inevitable to leverage statistical approaches in table detection.

To address the need for generalization, statistical machine learning approaches have been proposed to alleviate these problems. \cite{10.1007/3-540-48172-9_21} was one of the first to apply unsupervised learning method to the table detection task back in 1998. Their recognition process differs significantly from previous approaches as it realizes a bottom-up clustering of given word segments, whereas conventional table structure recognizers all rely on the detection of some separators such as delineation or significant white space to analyze a page from the top-down. In 2002, \cite{1047838} started to use supervised learning method by means of a hierarchical representation based on the MXY tree. The algorithm can be adapted to recognize tables with different features by maximizing the performance on an appropriate training set. After that, table detection has been cast into a set of different machine learning problems such as sequence labeling~\cite{Silva:2009:LRH:1634930.1635458}, feature engineering with SVM~\cite{6628801} and also ensemble a set of models~\cite{DBLP:journals/corr/FanK15} including Naive Bayes, logistic regression, and SVM. The application of machine learning methods has significantly improved table detection accuracy.

Recently, the rapid development of deep learning in computer vision has a profound impact on the data-driven image-based approaches for table detection. The advantage of the images-based table detection is two-fold: First, it is robust to document types by making no assumption of whether scanned images of pages or natively-digital document formats. Second, it reduces the efforts of hand-crafted feature engineering in conventional machine learning.~\cite{7490132} first used convolutional neural networks in table detection, where some table-like areas are selected first by some loose rules, and then the convolutional networks are built and refined to determine whether the selected areas are tables or not. In addition, the Faster R-CNN model has also been used in table detection.~\cite{8270123} presented a system that is totally data-driven and does not need any heuristics or metadata to detect as well as to recognize tabular structures on the ICDAR 2013 dataset. At the same time,~\cite{8270062} achieved state-of-the-art performance using the Faster R-CNN model on the UNLV dataset. Despite the success of applying deep learning models, most of these methods fine-tune pre-trained models on out-of-domain data with just a few thousand human-labeled examples, which is still difficult to be practicable in real-world applications. Till now, there is not any standard benchmark training dataset for both table detection and recognition. Therefore, we make the TableBank dataset publicly available and hope it can release the power of many deep learning approaches.

\subsection{Table Structure Recognition}

Table structure recognition aims to identify the row and column layout information for a table. The research of table structure recognition also includes rule-based approaches, machine learning approaches, and deep learning approaches. \cite{ramel2003detection} developed a method using the analysis of the graphic lines to detect and extract tables. After that, \cite{yildiz2005pdf2table} used the `pdftohtml\footnote{\small \url{http://pdftohtml.sourceforge.net/}}' tool that returns all text elements in a PDF file and computed horizontal overlaps of texts to recognize columns. \cite{hassan2007table} grouped tables into three categories by the principle if a table has horizontal or vertical ruling lines, and they developed heuristic methods to detect these tables. Unlike other methods designing rules for specified tables, \cite{shigarov2016configurable} presented a more general rule-based approach to apply to different domains. For machine learning-based approaches, \cite{doi:10.1117/12.410859} identified columns by hierarchical clustering and then spatial and lexical criteria to classify headers, and also addressed the evaluating problem by designing a new paradigm ``random graph probing''.

Recently, \cite{8270123} used the deep learning-based object detection model with pre-processing to recognize the row and column structures for the ICDAR 2013 dataset. Similarly, existing methods usually leveraged no training data or only small scale training data for this task. Distinct from existing research, we use the TableBank dataset to verify the data-driven end-to-end model for structure recognition. To the best of our knowledge, the TableBank dataset is the first large scale dataset for both table detection and recognition tasks.


\section{Conclusion}

To empower the research of table detection and structure recognition for document analysis, we introduce the TableBank dataset, a new image-based table analysis dataset built with online Word and Latex documents. We use the Faster R-CNN model and image-to-text model as the baseline to evaluate the performance of TableBank. In addition, we have also created testing data from Word and Latex documents respectively, where the model accuracy in different domains is evaluated. Experiments show that image-based table detection and recognition with deep learning is a promising research direction. We expect the TableBank dataset will release the power of deep learning in the table analysis task, meanwhile fosters more customized network structures to make substantial advances in this task.

For future research, we will further enlarge the TableBank from more domains with high quality. Moreover, we plan to build a dataset with multiple labels such as tables, figures, headings, subheadings, text blocks and more. In this way, we may obtain fine-grained models that can distinguish different parts of documents.

\section{Acknowledgments}
This work was supported in part by the National Natural Science Foundation of China (Grant Nos.U1636211, 61672081,61370126),the Beijing Advanced Innovation Center for Imaging Technology(Grant No.BAICIT-2016001), and the Fund of the State Key Laboratory of Software Development Environment (Grant No.SKLSDE-2019ZX-17).

\section{Bibliographical References}
\label{main:ref}
\bibliographystyle{lrec}
\bibliography{tablebank-ref}

\end{document}